\documentclass{llncs}
\usepackage{subcaption}
\usepackage{array,float}
\usepackage{amsfonts, amssymb}
\usepackage{graphicx,amsmath}
\usepackage{pgfplots}
\pgfplotsset{compat=1.8}
\usepgfplotslibrary{statistics}
\usepackage{tikz}
\usepackage{hyperref}
\usepackage{caption}
\usepackage{comment}
\usepackage{relsize}
\usepackage{soul}
\usepackage{tikz-3dplot}
\usepackage[titletoc, title]{appendix}
\usetikzlibrary{positioning,arrows}
\tikzstyle{place}=[circle,draw,text width=1.5em, text centered,minimum size=0.1cm]
\usetikzlibrary{arrows.meta}
\usepackage{makeidx}  % allows for indexgeneration
\usepackage{multicol}

\def \Em{{\mathbb{E}}}
\def \Rm{{\mathbb{R}}}

\def \Im{{\mathbb{I}}}
\def \abf{{\mathbf a}}

\def \sbf{{\mathbf s}}
\def \Sbf{{\mathbf S}}

\def \xbf{{\mathbf x}}
\def \Xbf{{\mathbf X}}

\def \0bf{{\mathbf 0}}

\newcommand{\alphabf}{\ensuremath{\boldsymbol{\mathlarger\alpha}}} 

\def \Ncal{{\mathcal N}}

\def \Fcal{{\mathcal F}}
\def \Ecal{{\mathcal E}}

\def \Qcal{{\mathcal Q}}
\def \Gcal{{\mathcal G}}

\usepackage{lineno}

\begin{document}
\mainmatter              % start of the contributions
\title{Mean Field Network based Graph Refinement with application to Airway Tree Extraction}
\titlerunning{MFA}  % abbreviated title (for running head)
%\begin{comment}
\author{Raghavendra Selvan\inst{1}, Max Welling\inst{2,3}, Jesper H. Pedersen\inst{4}, Jens Petersen\inst{1},
	Marleen de Bruijne\inst{1,5}}
%\author{***}
%
\authorrunning{} % abbreviated author list (for running head)
\tocauthor{}
\institute{***}
%\begin{comment}
\institute{Department of Computer Science, University of Copenhagen
\and
Informatics Institute, University of Amsterdam
\and
Canadian Institute for Advanced Research
\and
Department of Cardio-Thoracic Surgery RT, University Hospital of Copenhagen
\and
Departments of Medical Informatics and Radiology, Erasmus Medical Center
\\
\email{raghav@di.ku.dk}
}
%\end{comment}
\authorrunning{} % abbreviated author list (for running head)

\maketitle              % typeset the title of the contribution

%Unlike most segmentation methods used for tree extraction that are sequential, our method is exploratory in nature and is less sensitive to local anomalies in the data due to acquisition noise and/or interfering structures. 

\begin{abstract}
We present tree extraction in 3D images as a graph refinement task, of obtaining a subgraph from an over-complete input graph. To this end, we formulate an approximate Bayesian inference framework on undirected graphs using mean field approximation (MFA). Mean field networks are used for inference based on the interpretation that iterations of MFA can be seen as feed-forward operations in a neural network. This allows us to learn the model parameters from training data using back-propagation algorithm. We demonstrate usefulness of the model to extract airway trees from 3D chest CT data. We first obtain probability images using a voxel classifier that distinguishes airways from background and use Bayesian smoothing to model individual airway branches. This yields us joint Gaussian density estimates of position, orientation and scale as node features of the input graph. Performance of the method is compared with two methods: the first uses probability images from a trained voxel classifier with region growing, which is similar to one of the best performing methods at EXACT'09 airway challenge, and the second method is based on Bayesian smoothing on these probability images. Using centerline distance as error measure the presented method shows significant improvement compared to these two methods.
% * <marleen.de.bruijne@gmail.com> 2018-02-27T11:50:35.893Z:
% 
% >  two methods:
% can we argue these are state of the art (eg with reference to current EXACT results)? If so,  argue for it  in the paper and mention it here.
% 
% ^.
% * <marleen.de.bruijne@gmail.com> 2018-02-27T11:49:06.614Z:
% 
% > Gaussian density node features
% sounds ok, but unclearr what you mean. If you have space left clarify this point.
% 
% ^.
% * <marleen.de.bruijne@gmail.com> 2018-02-27T11:44:40.205Z:
% 
% > image segmentation based on super-voxel representations, specifically centerline extraction,
% 
% 
% image segmentation and centerline extraction are very different problems,  consequently phrases like " a segmentation method for centerline extraction" are confusing. I recommend to present it as a tree extraction approach and in the introduction and or discussion make the link to super voxel segmentation. 
% 
% For supervoxels, look at more recent work in the miccai community based on SLIC supervoxels to cite (I have seen papers based on SLUC+MRF, not sure if they are published though)
% 
% I would not call your blob representation a supervoxel representation. supervoxel is typically a small connected region in the image.
% 
% ^.

\end{abstract}

\section{Introduction}

Markov random field (MRF) based image segmentation methods have been successfully used in several medical image applications~\cite{segmentation2001Zhang,LearningCRF2014Orlando}. Pixel-level MRF's are commonly used for segmentation purposes to exploit the regular grid nature of images. These models become prohibitively expensive  when dealing with 3D images, which are commonly encountered in medical image analysis. However, there are classes of methods that work with supervoxel representation to reduce density of voxels by abstracting local information as node features~\cite{superpixel2009Wang,extraction2017Selvan}. Image segmentation, in such models, can be interpreted as connecting voxels/supervoxels to extract the desired structures of interest. This has similarities with performing graph refinement, where an over-complete input graph is processed to obtain a subgraph that corresponds to structures of interest.
%where the input graph is over-complete and a subgraph corresponds to the connectivity between voxels/supervoxels that belong to the structures of interest. From this perspective, tree extraction can also be interpreted as a graph refinement task.
%, with the objective of extracting a subgraph from an over-complete input graph that expresses connectivity in the desired tree.
% * <marleen.de.bruijne@gmail.com> 2018-02-27T18:37:03.843Z:
% 
% > From this perspective, centerline extraction
% in case you keep the more general intro starting from voxel/supervoxel segmentation, first indicate that such problems can be seen as graph refinement before indicating that centerline extraction can be seen as graph refinement (I guess this comments till falls under teh category centerline extraction is not segmentation)
% 
% clarify that you assume the original (input) graph is overcomplete and how that is a reasonable assumption
% 
% ^.
%This is an acceptable last paragraph introduction, but I would try to focus it a bit more on what is good about the method. My take on that would be something on it being exploratory, model based, yet can be very easily extended with more freely learned parameters.

In this work, we present a novel approach to tree extraction by formulating it as a graph refinement procedure on MRF using mean field networks (MFN). We recover a subgraph corresponding to the desired tree structure from an over-complete input graph either by retaining or removing edges between pairs of nodes. We use supervoxel-like representation to associate nodes in the graph with features that make the input graph sparser. 
%and recover the underlying connectivity between these nodes from an over-complete graph. centerline 
We formulate a probabilistic model based on unary and pairwise potential functions that capture nodes and their interactions. The inference is performed using mean field networks which implement mean field approximation (MFA)~\cite{ImprovingMFA1998Jaakkola} iterations as feed-forward operation in a neural network~\cite{meanfield2014Li}. The MFN interpretation enables us to learn the model parameters from training data using back-propagation algorithm; this allows our model to be seen as an intermediate between entirely model-based and end-to-end learning based approaches. The proposed model is exploratory in nature and, hence, not sensitive to local anomalies in data. 
%While the interactions between the nodes are modeled, there is enough flexibility in the overall model to learn parameters from data. 
We evaluate the method to extract airway trees in comparison with two methods: the first uses probability images from a trained voxel classifier with region growing~\cite{vessel2010Lo}, which is similar to one of the best performing methods in EXACT airway challenge~\cite{extraction2012Lo}, and the second method is based on Bayesian smoothing on probability images obtained from the voxel classifier~\cite{extraction2017Selvan}. 
% * <marleen.de.bruijne@gmail.com> 2018-02-27T18:42:35.465Z:
% 
% > recover the underlying centerline
% clarify that you find which nodes are connected and which are not.
% 
% ^.
% * <marleen.de.bruijne@gmail.com> 2018-02-27T18:41:02.751Z:
% 
% > with a focus on extracting airway trees
% I'd move this a little later and state you evaluate this on airway extraction
% 
% ^.

\vspace{-0.1cm}

\section{Method}
\label{sec:method}
%\vspace{-0.2cm}
\subsection{The Graph Refinement Model}
\label{subsec:model}
	Given a fully connected, or over-complete, input graph, $\Gcal:\{\Ncal,\Ecal\}$ with nodes $i\in\Ncal$ and edges in $(i,j)\in\Ecal$, we are interested in obtaining a subgraph, ${\Gcal^{\prime}}:\{\Ncal^{\prime},\Ecal^{\prime}\}$, that in turn corresponds to a structure of interest like vessels or airways in an image. We assume each node $i\in\Ncal$ to be associated with a set of $d$-dimensional features, $\xbf_i \in \Rm^d$, and collected into a random vector, $\Xbf = [\xbf_1,\dots,\xbf_N]$.  We introduce a random variable, $\Sbf = [ \sbf_1 \dots \sbf_N]$,  to capture edge connections between nodes. Each node connectivity variable, $\sbf_i =\{s_{ij}\} : j = 1 \dots N$, is a collection of binary random variables, $s_{ij} \in \{0,1\}$, indicating absence or presence of an edge between nodes $i$ and $j$ and we are interested in recovering $\Sbf^\prime$ that describes the desired subgraph~$\Gcal^{\prime}$. Note that each instance of $\Sbf$ can be seen as an $N\times N$ adjacency matrix.
    
%[{\bf MAX: I would always just talk about $P(S|X)$, i.e. the conditional distribiution, and not P(X,S)}]    
The model described by the conditional distribution, $p(\Sbf|\Xbf)$, bears similarities with hidden MRF models that have been used for image segmentation\cite{segmentation2001Zhang,LearningCRF2014Orlando}. 
%In this terminology, the observed feature vector, $\Xbf$, is the observable random field and the binary connectivity variable, $\Sbf$, can be seen as the hidden label field with binary label classes. 
Based on this connection, we use the notion of node, $\phi_{i} (\sbf_i)$, and pairwise, $\phi_{ij}(\sbf_i,\sbf_j)$,  potentials to write the logarithm of joint distribution and relate it to the conditional distribution as,
	\begin{align}
	\ln p(\Sbf| \Xbf) \propto \ln p(\Sbf, \Xbf) =
		-\ln Z+\sum_{i \in \Ncal} \phi_{i} (\sbf_i) + \sum_{(i,j) \in \Ecal} \phi_{ij}(\sbf_i,\sbf_j), 
    \label{eq:cliq}
\end{align}
%[{\bf MAX: Don't forget the normalization constant, so it's better to use $\propto$ rather than $=$ (or explicitly write the $\log(Z)$)}]
where $\ln Z$ is the normalisation constant. For ease of notation, explicit dependence on observed data in these potentials is not shown. 

Next we focus on formulating node and pairwise potentials introduced in~\eqref{eq:cliq} to reflect the behaviour of nodes and their interactions in the subgraph $\Gcal^\prime$, which can consequently 
%By designing potentials that can best capture the node behaviours and their interactions in $\Gcal^\prime$, we can
yield good estimates of $p(\Sbf|\Xbf)$. First, we propose a node potential that imposes a prior degree on each node and learns a per-node feature representation that can be relevant to nodes in the underlying subgraph, $\Gcal^{\prime}$.
%We design parameterised potentials that contribute positively to the ELBO in~\eqref{eq:elbo} when a particular node or pairwise interaction is to be encouraged and negatively when it must be discouraged. 
For each node $i\in\Ncal$, it is given as, 
\begin{equation}
	\phi_i(\sbf_i) = \sum_{v=0}^{D} \beta_v \Im \Big [ \sum_{j} s_{ij} = v\Big ] +  \abf^T \xbf_i\sum_{j} s_{ij},
	\label{eq:phiN}
\end{equation}
where $\sum_j s_{ij}$ is the degree of node $i$ and $\Im[\cdot]$ is the indicator function. The parameters $\beta_v \in \Rm,\text{ } \forall \text{ } v = [0,\dots,D]$, can be seen as a prior on the degree per node. We explicitly model and learn this term for upto 2 edges per node and assume uniform prior for $D>2$. Further, individual node features, $\xbf_i$, are combined with $\abf\in \Rm^{d \times 1} $ and captures a combined node feature representation that is characteristic to the desired subgraph $\Gcal^\prime$. The degree of each node, $\sum_{j} s_{ij}$, controls the extent of each node's contribution to the node potential. 
%this is fur degree of the node $\sum_{j} s_{ij}$, controls the extent of contribution to the node potential its contribution to the node potential  and contributes to the node potential depending on its degree $\sum_{j} s_{ij}$. 

Secondly, we model the pairwise potential such that it captures interactions between pairs of nodes and is crucial in deciding the existence of edges between nodes. We propose a potential that enforces symmetry in connections, and also has terms that derive joint features for each pair of nodes that are relevant in prediction of edges, and is given as,
\begin{equation}
	\phi_{ij}(\sbf_i,\sbf_j) = \lambda \big( 1-2|s_{ij} - s_{ji}| \big ) + (2s_{ij}s_{ji}-1) \Big [ \boldsymbol{\eta}^T|\xbf_i-\xbf_j| + \boldsymbol{\nu}^T(\xbf_i\xbf_j)\Big]. 
	\label{eq:phiE}
\end{equation}
The function parameterised by $\lambda \in \Rm$ in~\eqref{eq:phiE} ensures symmetry in connections between nodes, i.e, for nodes $i,j$ it encourages $s_{ij} = s_{ji}$. The parameter $\boldsymbol{\eta} \in \Rm^{d \times 1}$  combines the absolute difference between each feature dimension. The element-wise feature product term $(\xbf_i\xbf_j)$ with $\boldsymbol{\nu} \in \Rm^{d \times 1}$ is a weighted, non-stationary polynomial kernel of degree 1 that computes the dot product of node features in a weighted feature space. %This becomes evident when the term is written as $k(\xbf_i,\xbf_j) = (\xbf_i^T\Wbf)^T (\xbf_j^T\Wbf) =\boldsymbol{\nu}^T(\xbf_i\xbf_j) $ where $\Wbf$ is a diagonal matrix with $\boldsymbol{\nu}^{1/2}$ as its diagonal elements ~\cite{weighted2016Wei}. 

Under these assumptions, the posterior distribution, $p(\Sbf|\Xbf)$, can be used to extract the subgraph, ${\Gcal^{\prime}}$ from $\Gcal$. However, except for in trivial cases, it is intractable to estimate $p(\Sbf|\Xbf)$ and we must resort to making some approximations. We take up the variational mean field approximation (MFA)~\cite{ImprovingMFA1998Jaakkola}, which is a structured approach to approximating $p(\Sbf|\Xbf)$ with candidates from a class of simpler distributions$: q(\Sbf) \in \Qcal$. This approximation is performed by minimizing the exclusive Kullback-Leibler divergence~\cite{ImprovingMFA1998Jaakkola}, or equivalently maximising the evidence lower bound (ELBO) or variational free energy, given as
\begin{equation}
 \Fcal(q_\Sbf) = \ln Z+ \Em_{q_{\Sbf}} \Big [ \ln p(\Sbf| \Xbf) - \ln q(\Sbf) \Big ],
 \label{eq:elbo}
\end{equation}
%[{\bf MAX: Also switch to $P(S|X)$ here}]
where $\Em_{q_{\Sbf}}$ is the expectation with respect to the distribution ${q_{\Sbf}}$. In MFA, the class of distributions, $\Qcal$, are constrained such that $q(\Sbf)$ can be factored further. In our model, we assume the existence of each edge is independent of the others, which is enforced as the following factorisation: 
%for distributions in~$\Qcal$,
\begin{equation}
q(\Sbf) = \prod_{i=1}^N \prod_{j=1}^N q_{ij}(s_{ij}),  \text{ where }
q_{ij}(s_{ij}) = \begin{cases}
\alpha_{ij} &\qquad \text{if }s_{ij} = 1 \\
(1-\alpha_{ij}) &\qquad \text{if }s_{ij} = 0
\end{cases},
\label{eq:mfa}
\end{equation}
where $\alpha_{ij} \in [0,1]$ is the probability of an edge existing between nodes $i$ and $j$.

Using the potentials from~\eqref{eq:phiN} and~\eqref{eq:phiE} in~\eqref{eq:elbo} and taking expectation with respect to the distribution $q_\Sbf$, we obtain the ELBO in terms of~$\alpha_{ij} \text{ }\forall\text{ } i,j=[1,\dots,N]$, proof of which is shown in the Appendix~\ref{sec:app}.
\begin{comment}
as,
\begin{align}	
	& \Fcal{q_{\Sbf}} =   \sum_{i \in \Ncal} \prod_{j \in \Ncal_i} (1-\alpha_{ij}) \Big \{ \beta_0  
+ \sum_{m \in \Ncal_i} \frac{\alpha_{im}} {(1-\alpha_{im})} \Big[ \beta_1   + \beta_2 \sum_{n \in \Ncal_i \setminus m} \frac{ \alpha_{in}}{(1-\alpha_{in})} \Big] 
\nonumber \\
&+
\abf^T \xbf_i\sum_{j} \alpha_{ij} \Big\}
 + \sum_{i \in \Ncal} \sum_{j \in \Ncal_i} \Big\{ \lambda \big( 1-2(\alpha_{ij} +\alpha_{ji}) +4\alpha_{ij}\alpha_{ji}\big) -\Big( \alpha_{ij} \ln {\alpha_{ij}} \nonumber \\
 &  + (1-\alpha_{ij}) \ln (1-{\alpha_{ij}}) \Big) 
 + (2\alpha_{ij}\alpha_{ji}-1) \Big [ \boldsymbol{\eta}^T|\xbf_i-\xbf_j| + \boldsymbol{\nu}^T(\xbf_i\xbf_j)\Big]\Big \}.
	\label{eq:elbo_aw}
\end{align}
\end{comment}
By differentiating this ELBO with respect to any individual $\alpha_{kl}$, 
%in effect with respect to the approximating factor $q_{kl}(s_{kl})$, 
we obtain the following update equation for performing MFA iterations. At iteration $(t+1)$: 
\begin{equation}
	\mathlarger \alpha_{kl}^{(t+1)} = \mathlarger \sigma({\gamma_{kl}}) = \frac{1}{1+\exp^{-\gamma_{kl}}} \text{ } \forall \text{ } {k} = \{1\dots N\},\text{ } l \in \Ncal_k \text{ }:\text{ } |\Ncal_k|=L
    \label{eq:mfaUp}
\end{equation}
where $\mathlarger \sigma(\cdot)$ is the sigmoid activation function, $\Ncal_k$ are the $L$ nearest neighbours of node $k$ based on positional Euclidean distance, and 
\begin{align}
	&\mathlarger \gamma_{kl} = 
\prod_{j \in \Ncal_k \setminus l} \big(1-\alpha_{kj}^{(t)}\big) \Big\{ \sum_{m \in \Ncal_k \setminus l} \frac{\alpha_{km}^{(t)}}{(1-\alpha_{km}^{(t)})}
	\Big[ (\beta_2-\beta_1) - \beta_2 \sum_{n \in \Ncal_k \setminus l,m} \frac{\alpha_{kn}^{(t)}}{(1-\alpha_{kn}^{(t)})}\Big]
     \nonumber \\ 
    &+ \big(\beta_1-\beta_0 \big) \Big\} 
    + \abf^T \xbf_i + (4\alpha_{lk}^{(t)}-2)\lambda + 2\alpha_{lk}^{(t)}\big( \boldsymbol{\eta}^T|\xbf_i-\xbf_j| + \boldsymbol{\nu}^T(\xbf_i\xbf_j) \big).
    \label{eq:mfaNN}
\end{align}
After each iteration $t$, MFA outputs $N\times N$ edge predictions, which we denote as ${\alphabf}^{(t)}$, with entries $\mathlarger\alpha_{kl}^{(t)}$. MFA iterations are performed until convergence, and a good stopping criteria is when the increase in ELBO is below a small threshold between successive iterations. Note that an estimate of the connectivity variable $\Sbf$ at iteration $t$ can be recovered as $\Sbf^{(t)} = \Im[{\alphabf}^{(t)} > 0.5]$.

%Since an estimate of the random variable at iteration $t$ can be recovered as, $\Sbf^{(t)} = \Im[{\alphabf}^{(t)} > 0.5]$, it fully describes the approximating distribution $q^{(t)}{(\Sbf)}$.
%[{\bf MAX: This is confusing, please do not confuse the random variable S with the probability $\alpha$. You can recover an estimate of S by picking the state: $S^* = I[\alpha > 0.5]$}]
\vspace{-0.1cm}
\subsection{Mean Field Network}
The MFA update equations in~\eqref{eq:mfaUp} and~\eqref{eq:mfaNN} resemble the computations in a feed-forward neural network. The predictions from iteration $t$, ${\alphabf}^{(t)}$, are combined and passed through a non-linear activation function, a sigmoid in our case, to obtain predictions at iteration $t+1$, ${\alphabf}^{(t+1)}$. 
%[{\bf MAX: Please use the probabilities $\alpha$ as the hidden variables inthe MFN as the states S are discrete.}] 
This interpretation can be used to map $T$ iterations of MFA to a $T$-layered neural network, based on the underlying graphical model, and is seen as the mean field network (MFN)~\cite{meanfield2014Li}. The parameters of our model form weights of such a network and are shared across all layers. Given this setting, parameters for the MFN model can be learned using back-propagation on the binary cross entropy (BCE) loss computed as,
	\begin{equation}
		\mathcal{L}(\Sbf^\prime, {\alphabf}^{(T)}) = \frac{1}{N^2}  \sum_{i=1}^N\sum_{j=1}^N \Big( {s}_{ij}\log(\alpha_{ij}) + (1-{s}_{ij})\log(1-\alpha_{ij})\Big),
        \label{eq:bce}
	\end{equation}
    %[{\bf MAX: you already wrote the ELBO, so you can remove this equation now}]
where ${\alphabf}^{(T)}$ is the predicted probability of edge connections at the last iteration $(T$) of MFA and ${\Sbf^\prime}$ is the ground truth adjacency of the desired subgraph $\Gcal^\prime$. 
%[{\bf MAX: You do not explain anywhere with an equation that you are actually backprogating on the error between the predictions from your model and the ground truth connections. I think it's important you make that expolicit. You could replace the equation above with the equation for the cross entropy loss or L2 loss (whatever you are using).}]
\begin{comment}

\end{}
\begin{figure}
\centering
\includegraphics[height=0.5\linewidth,width=0.75\linewidth]{figures/mfn.pdf}
\caption{Overview of the graph-refinement procedure proposed using mean field networks. An over-complete graph is input to the network at layer-$(1)$ with initial connection probabilities $\alphabf^{(0)}$ and trained using back-propagation to output the subgraph, in this case, airway tree.}
\label{fig:mfn}
\end{figure}
\end{comment}
\vspace{-0.1cm}
\subsection{Airway Tree Extraction as Graph Refinement}

Depending on the input features of observed data, $\Xbf$, the MFN presented above can be applied to different applications. Here we present extraction of airway tree centerlines from CT images as a graph refinement task 
%as illustrated in the overview in Figure~\ref{fig:mfn} 
and show related experiments in Section~\ref{sec:res}. To this end, the image data is processed to extract useful node features to input to the MFN. We assume that each node is associated with a 7-dimensional Gaussian density comprising of location ($x,y,z$), local radius ($r$), and orientation $(v_x,v_y,v_z)$, such that $\xbf_i = [\xbf^i_{\mu}, \xbf^i_{\sigma^2}]$, comprising of mean, $\xbf^i_{\mu}\in \Rm^{7\times 1}$, and variance for each feature, $\xbf^i_{\sigma^2}\in \Rm^{7\times 1}$. We obtain these features by performing Bayesian smoothing on probability images obtained from the voxel classifier~\cite{vessel2010Lo}, with process and measurement models that model individual branches in an airway tree using the method of~\cite{extraction2017Selvan}. 
% * <marleen.de.bruijne@gmail.com> 2018-02-27T19:27:44.520Z:
% 
% >  the image data
% clarify it is not raw image data but probability map from a classifier
% 
% ^.

The node and pairwise potentials in equations~\eqref{eq:phiN} and~\eqref{eq:phiE} are general and applicable to commonly encountered trees. The one modification we make due to our feature-based representation is to one of the terms in~\eqref{eq:phiE}, where we normalise the absolute difference in node positions, $\xbf_p = [x,y,z]$, with the average radius of the two nodes, i.e.,  $|\xbf_p^i-\xbf_p^j|/(r^i+r^j)$, as the relative positions of nodes are proportional to their scales in the image.

For evaluation purposes, we convert the refined graphs into binary segmentations by drawing spheres in 3D volume along the predicted edges using location and scale information from the corresponding node features.  

% * <marleen.de.bruijne@gmail.com> 2018-02-27T19:29:04.031Z:
% 
% > that is specific to extraction of airways 
% it is not specific to the extraction of airways, rather to the sparse blob representation you choose (could work the same for e.g. vessel structures). 
% 
% ^.
\vspace{-0.1cm}
\section{Experiments and Results}
\label{sec:res}
%\vspace{-0.2cm}
\vspace{-0.1cm}
\subsubsection{Data }
The experiments were performed on 32 low-dose CT chest scans from a lung cancer screening trial~\cite{danish2009Pedersen}. All scans have  voxel-resolution of approximately  $0.78\times 0.78 \times 1$mm$^3$. The reference segmentations consist of expert-user verified union of results from two previous methods: first method uses a voxel classifier to distinguish airway voxels from background to obtain probability images and extracts airways using region growing and vessel similarity~\cite{vessel2010Lo}, and the second method continually extends locally optimal paths using costs computed using the voxel classification approach~\cite{airway2009Lo}. We extract ground truth adjacency matrices for training the MFN using Bayesian smoothing to extract individual branches from the probability images obtained using the voxel classifier, then connect only the branches within the reference segmentation to obtain a single, connected tree structure using a spanning-tree algorithm.

% * <marleen.de.bruijne@gmail.com> 2018-02-27T19:38:50.955Z:
% 
% >  them 
% into a single, connected tree structure?
% 
% ^.
% * <marleen.de.bruijne@gmail.com> 2018-02-27T19:36:46.650Z:
% 
% > two previous methods
% clarify that both use a similar voxel classificaiton probability map as you do?
% 
% ^.
% * <marleen.de.bruijne@gmail.com> 2018-02-27T19:35:57.412Z:
% 
% >  first method uses a voxel classifier that incorporates vessel orientation similarity in its appearance model to distinguish airway voxels from background 
% the orientation is not used  in the classifier but in the region growing
% 
% ^.
% * <marleen.de.bruijne@gmail.com> 2018-02-27T19:30:49.619Z:
% 
% > we use 16 scans for training and 16 for test purposes.
% consider 4 or 8fold crossvalidation
% 
% ^.
\vspace{-0.2cm}
\subsubsection{Error Measure }

To evaluate the proposed method along with the comparison methods in a consistent manner, we extract centerlines using a 3D-thinning algorithm from the generated binary segmentations.
%To evaluate the extracted subgraphs from the MFN we use the binary segmentations, generated using the connectivity information in the predicted subgraphs and node features, to extract centerlines using a 3D-thinning algorithm. 
The error measure used is based on centerline distance, defined as $d_{err} = (d_{FP} + d_{FN})/2 $, where $d_{FP}$ is average minimum Euclidean distance from segmented centerline points to reference centerline points and captures false positive error, and $d_{FN}$ is average minimum Euclidean distance from reference centerline points to segmentation centerline points and captures false negative error.
% * <marleen.de.bruijne@gmail.com> 2018-02-27T19:41:01.445Z:
% 
% > generate 3D binary segmentations
% consider making this part of the (blob representation  specific) method instead of the evaluation only. Then it becomes logical to visualize the 3D segmentations instead of the centerlines. (Lung analysis readers will probably be interested in thsoe.)
% 
% ^.

\begin{figure}[ht!]
\centering
\begin{minipage}{0.48\textwidth}
\centering
\includegraphics[width=0.9\linewidth]{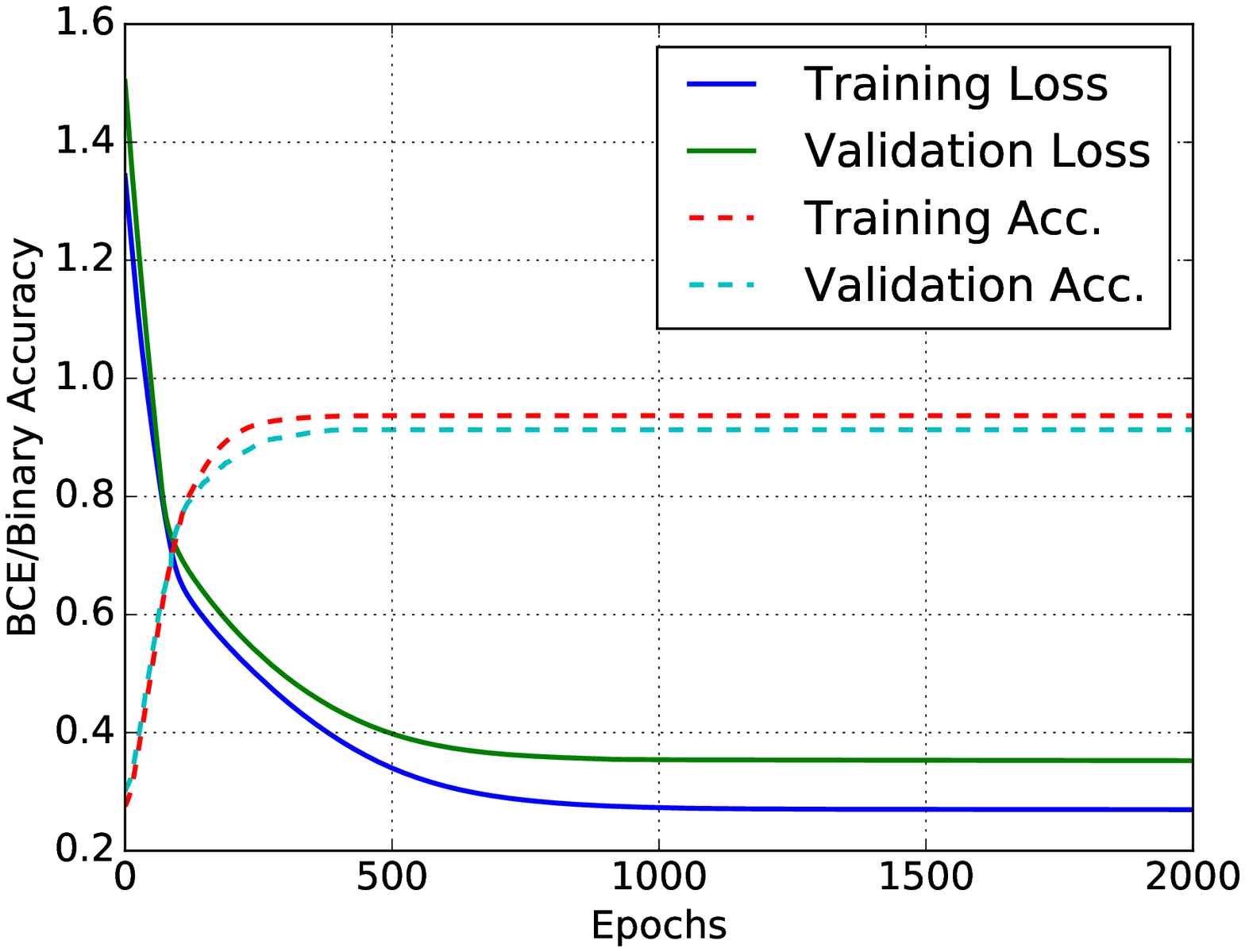}
\end{minipage}
\begin{minipage}{0.48\textwidth}
\centering
\includegraphics[width=0.9\linewidth]{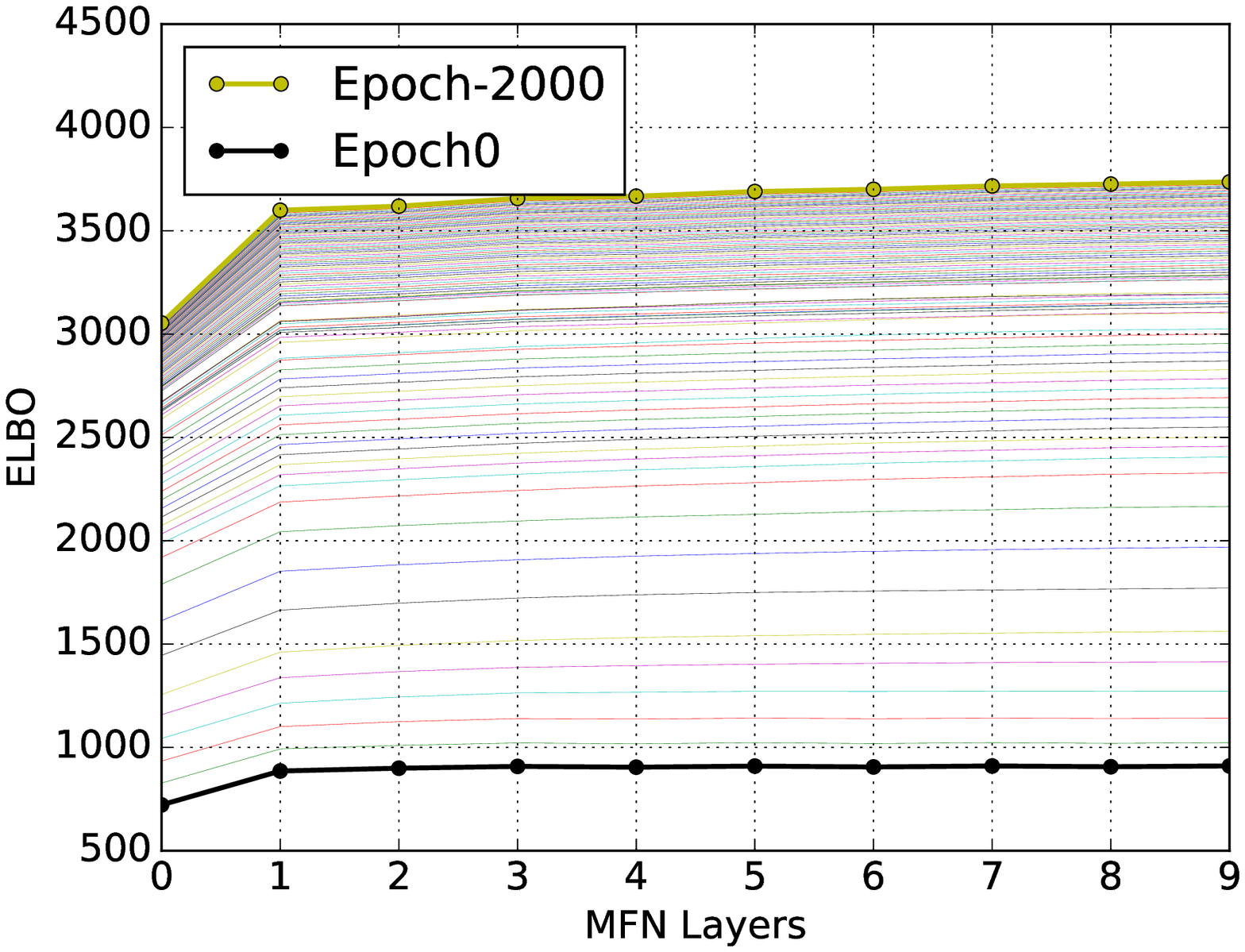}
\end{minipage}
\caption{Training and validation losses for MFN along with the binary accuracies averaged over the four-folds of the cross-validation procedure are shown in the figure to left. Binary accuracy is obtained by thresholding predicted probability, i.e., $\Sbf^{(t)} = \Im[{\alphabf}^{(t)} > 0.5]$. Figure to the right shows the ELBO computed at each layer within an epoch and across epochs averaged over four folds.}
\label{fig:learning}
\vspace{-0.5cm}
\end{figure}
\vspace{-0.1cm}
\subsubsection{Learning of Parameters }

%We perform four-fold cross validation to evaluate our method. Images in the training folds are used to train the MFN and tested on the ones in validation fold for each run of cross validation. 
We create sub-images comprising of $500$ nodes (batch) from each image and derive the corresponding adjacency matrices to reduce memory footprint during the training procedure. Parameters of the MFN are learned by minimizing the BCE loss in~\eqref{eq:bce} computed using all batches of all images in the training data using back-propagation with Adam optimiser with recommended settings~\cite{adam2014Kingma}. To further reduce computational overhead we restrict the neighbourhood of each node to be $L=10$ nearest neighbours based on Euclidean distance of their locations in the image data. Based on initial investigations of ELBO we set the number of layers in MFN, $T=10$. %The MFN is implemented in PyTorch with a custom forward layer implementing the MFA update in equations~\eqref{eq:mfaUp} and~\eqref{eq:mfa}. 
The learning curves for loss and binary accuracy are shown in Figure~\ref{fig:learning}, along with the ELBO plot showing the successive increase in ELBO with each iteration within an epoch (as guaranteed by MFA) and with increasing epochs (due to gradient descent). 

% * <marleen.de.bruijne@gmail.com> 2018-02-27T19:51:53.393Z:
% 
% > reduce memory footprint 
% so is there only these 500 nodesl in one batch for optimization or do you compute gradient on batches of these batches?
% 
% ^.
% * <marleen.de.bruijne@gmail.com> 2018-02-27T19:50:33.924Z:
% 
% > We create batches of $500$ nodes from each image
% how? (I assume subimages but will be good to specify)
% 
% ^.
%\newcolumntype{K}[1]{>{\centering\arraybackslash}p{#1}}
\begin{table*}[h]
\vspace{-0.5cm}
\caption{ Performance comparison based on 4-fold cross validation.}
% * <marleen.de.bruijne@gmail.com> 2018-02-27T19:53:59.838Z:
% 
% > \caption{ Performance comparison on the test set}
% add statistical test to test for statistical significance of differences
% 
% ^.
\label{tab:res}
\begin{center}
%\small{
    %\begin{tabular}{| l | K{2cm} | K{1.75cm}|  K{1.75cm}| K{2.25cm}|}
    \begin{tabular}{ lcccc}
    \hline
    Method & $d_{FP}$(mm) & $d_{FN}$(mm) & $d_{err}$ (mm) \\ \hline
    \small{Voxel Classifier} & $0.792$ & $4.807$ & $2.799 \pm 0.701 $ \\ %\hline
    \small{Bayesian Smoothing} & $0.839$ & $2.812$ & $1.825 \pm 0.232$  \\ %\hline
%    (RTS+RG)$_2$ & $0.401$ & $2.658$ & $1.529$ & $0.165$  \\ \hline
   % \textbf{MFN} &  $2.967$ & $1.173$ & $2.071$ & $0.179$  \\ \hline
    \textbf{MFN} &  $0.835$ & $2.571$ & $1.703 \pm 0.186$   \\ \hline
    \end{tabular}
\end{center}
\vspace{-1.5cm}
\end{table*}

\begin{figure}[ht!]
\centering
\begin{comment}
%\begin{minipage}{0.49\textwidth}
\begin{minipage}{0.24\textwidth}
\includegraphics[width=0.9\linewidth]{figures/vol1580_con.png}
\end{minipage}
\begin{minipage}{0.24\textwidth}
\includegraphics[width=0.9\linewidth]{figures/vol2287_con.png}
\end{minipage}
\begin{minipage}{0.24\textwidth}
\includegraphics[width=0.9\linewidth]{figures/vol1032_con.png}\end{minipage}
\begin{minipage}{0.24\textwidth}
\includegraphics[width=0.9\linewidth]{figures/vol1389_con.png}
\end{minipage}
\end{comment}
\begin{minipage}{0.49\textwidth}
\includegraphics[width=0.75\linewidth]{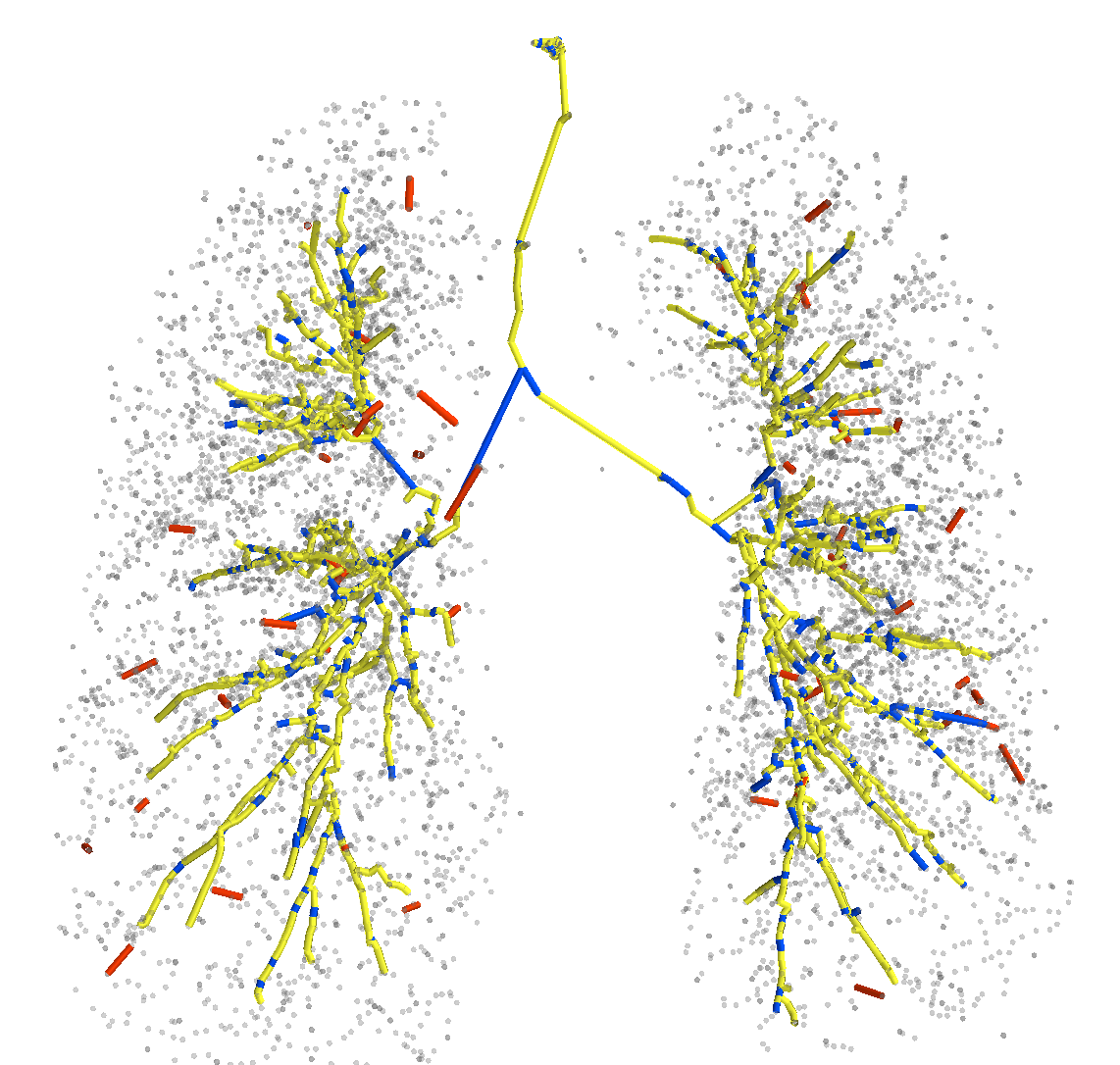}
\end{minipage}
\begin{minipage}{0.49\textwidth}
\begin{minipage}{0.49\textwidth}
\includegraphics[width=0.99\linewidth]{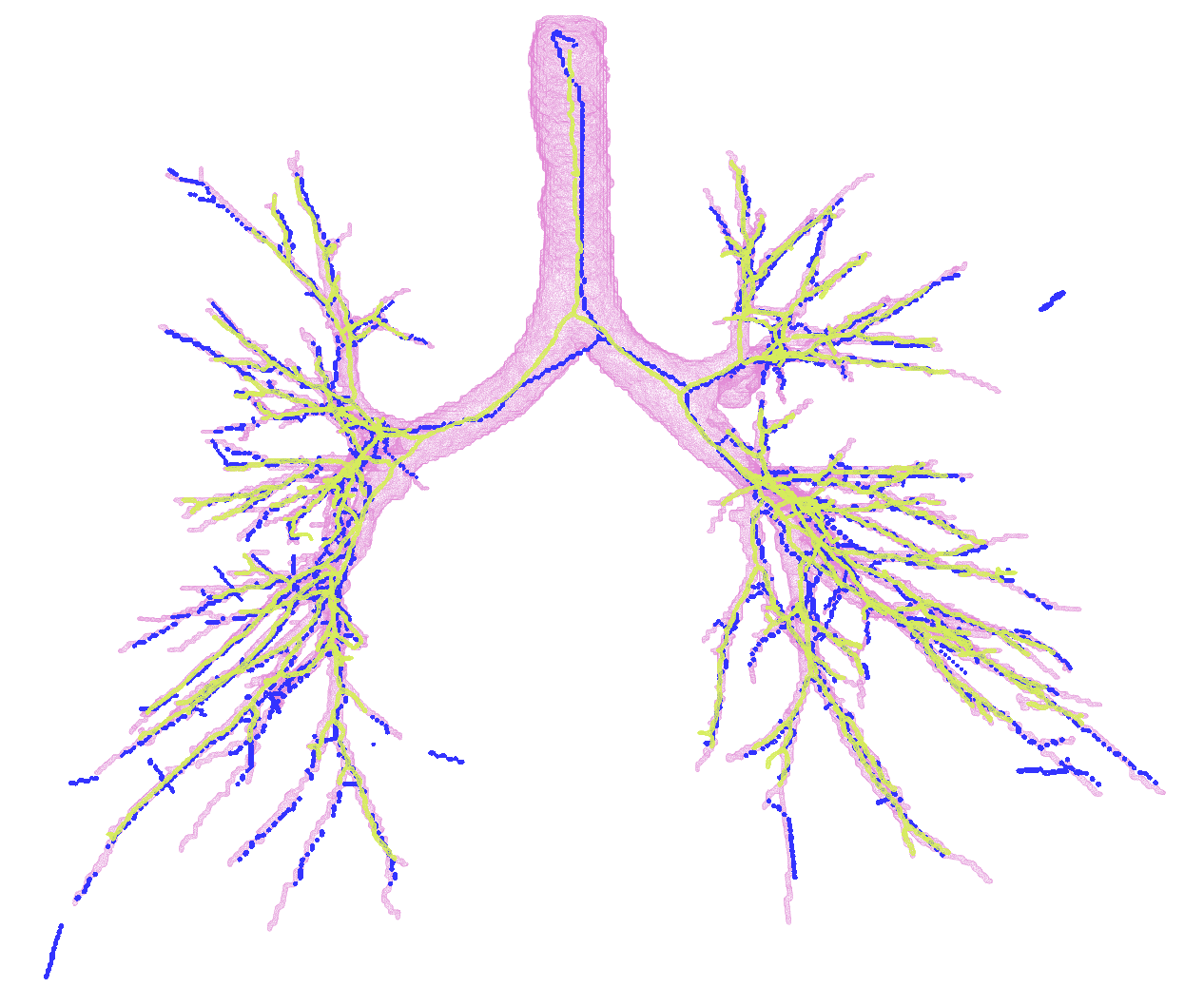}
\end{minipage}
\begin{minipage}{0.49\textwidth}
\includegraphics[width=0.99\linewidth]{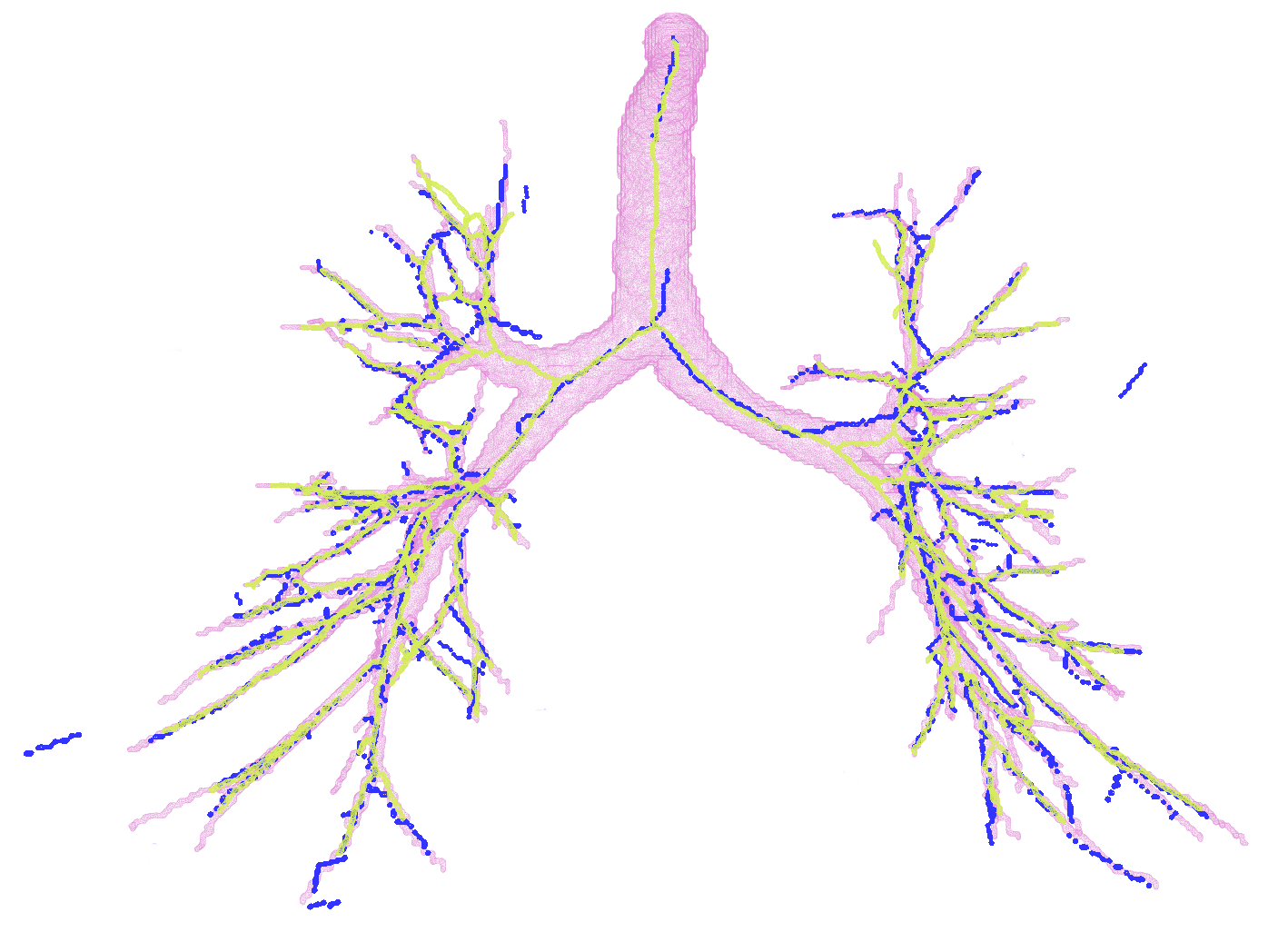}
\end{minipage}

\begin{minipage}{0.49\textwidth}
\includegraphics[width=0.99\linewidth]{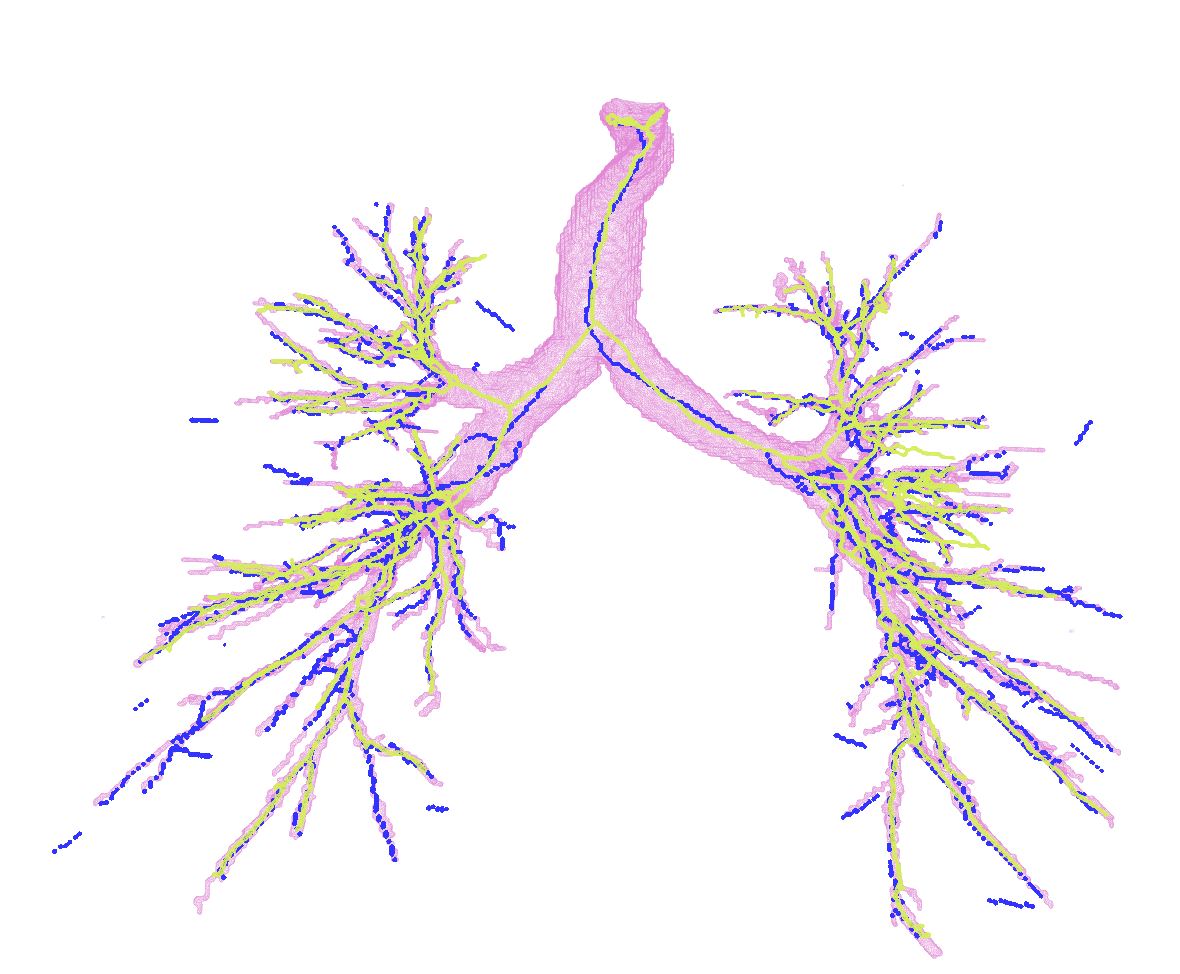}
\end{minipage}
\begin{minipage}{0.49\textwidth}
\includegraphics[width=0.99\linewidth]{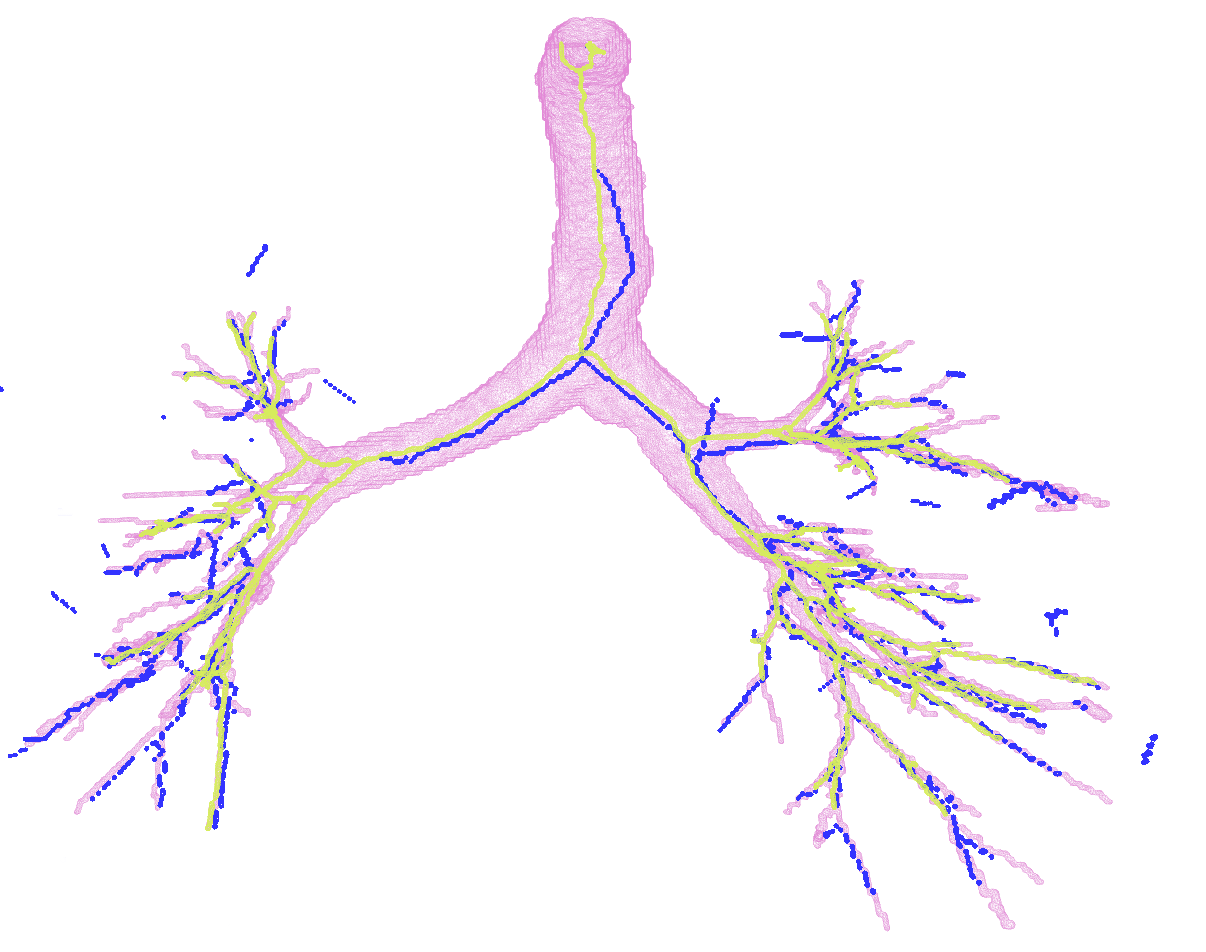}
\end{minipage}
\end{minipage}
\caption{Figure on left: Predicted connections by MFN for one case: Yellow edges are true positives, red edges are false positives and blue are false negatives. Figure on right: Airway tree centerlines for four cases obtained from MFN predictions (blue) overlaid with the reference segmentations (pink surface) and the centerlines from the voxel-classifier based region growing method (yellow). }
\label{fig:results}
\vspace{-0.5cm}
\end{figure}

%\vspace{-0.2cm}
\subsubsection{Results }

%The trained MFN was used to predict edge connectivity, $\tilde{\Sbf}^{(T)}$, on the test data  for each image. We derive a binary adjacency matrix by thresholding the edge probabilities in $\tilde{\Sbf}^{(T)}$ at $0.5$. Binary volume segmentations are generated from these binary adjacency predictions and the corresponding node features to obtain the centerlines. 
We compare performance of the proposed MFN method with a method close to the voxel classifier approach that uses region growing on probability images %with a threshold on probability of $0.5$ obtained using training
~\cite{vessel2010Lo} which was one of the best performing methods in the EXACT'09 challenge~\cite{extraction2012Lo}, and Bayesian smoothing method used in tandem with the voxel classifier approach~\cite{extraction2017Selvan}.  We perform 4-fold cross validation using the 32 images on all three methods and report centerline distance based performance measure, $d_{err}$ in mm, in Table~\ref{tab:res} based on the cross validation predictions.
% * <marleen.de.bruijne@gmail.com> 2018-02-27T19:59:39.991Z:
% 
% >  region growing on probability images with a threshold on probability of $0.5$
% 
% is this a fair comparison? The probability maps were not necessarily normalized (sampling was heavily biased towards foreground) and  the treshold was meant to be tuned.
% 
% ^.
% * <marleen.de.bruijne@gmail.com> 2018-02-27T19:56:28.986Z:
% 
% > Binary volume segmentations are generated from these binary adjacency predictions and the corresponding node features to obtain the centerlines. 
% here is slight overlap with  the error metric description.
% 
% ^ <marleen.de.bruijne@gmail.com> 2018-02-27T20:06:05.270Z:
% 
% ok, I thought this was part of the text but I now see it is not.  But then I miss an explanation of the baseline methods.
%
% ^.
Our method shows an improvement in the average error with significant gains $(p < 0.05)$ by reducing the false negative error $d_{FN}$, implying extraction of more complete trees, when compared to both methods. The results were compared based on paired-sample $t$-test.

In Figure~\ref{fig:results}, first we present the predicted subgraph for one of the images. The gray dots are nodes of the over-complete graph with features, $\xbf_i$, extracted using Bayesian smoothing; the edges are colour-coded providing an insight into the performance of the method: yellow edges are true positives, red edges are false positives and blue edges are false negatives compared to the ground truth connectivity derived from the reference segmentations. Several of the false negatives are spaced closely, and in fact, do not contribute to the false negative error, $d_{FN}$, after generating the binary segmentations.
% * <marleen.de.bruijne@gmail.com> 2018-02-27T20:06:46.156Z:
% 
% > by reducing the false negative error $d_{FN}$. 
% If you have time and space left, it would be interesting to investigate where these differences are - e.g. divide ground truth in generations and check where FN is reduced most
% 
% ^.
The figure to the right in Figure~\ref{fig:results} shows four predicted centerlines overlaid with the reference segmentation and centerlines from the voxel-classifier approach. Clearly, the MFN method is able to detect more branches as seen in most of the branch ends, which is also captured as the reduction in $d_{FN}$ in Table~\ref{tab:res}. Some of the false positive predictions from MFN method appear to be a missing branch in the reference as seen in the first of the four scans. However, there are few other false positive predictions that could be due to the model using only pairwise potentials; this can be alleviated either  by using higher order neighbourhood information or with basic post-processing. The centerlines extracted from MFN are slightly offset from the center of airways at larger scales; this could be due to the sparsity of the nodes at those scales and can be overcome by increasing resolution of the input graph.

\vspace{-0.1cm}
\section{Discussion and Conclusion}
\label{sec:disc}
\vspace{-0.1cm}
We presented a novel method to perform tree extraction by posing it as a graph refinement task in a probabilistic graphical model setting. We performed approximate probabilistic inference on this model, to obtain a subgraph representing airway-like structures from an over-complete graph, using mean field approximation. 
%from an over-complete graph representing all airway-like structures in the image,  the underlying subgraph representing the airway tree,  using the mean field approximation. 
Further, using mean field networks we showed the possibility of learning parameters of the underlying graphical model from training data using back-propagation algorithm. The main contribution within the presented MFN framework is our formulation of unary and pairwise potentials as presented in~\eqref{eq:phiN} and~\eqref{eq:phiE}. By designing these potentials to reflect the nature of tasks we are interested in, the model can be applied to a diverse set of applications. We have shown its application to extract airway trees with significant improvement in the error measure, when compared to the two comparison methods. However, tasks like tree extraction can benefit from using higher order potentials that take more than two nodes jointly into account. This limitation is revealed in Figure~\ref{fig:results}, where the resulting subgraph from MFN is not a single, connected tree. %Adding more expressive potentials that can capture this behaviour will make the model further complete. 
%These potentials also enabled learning of relevant features from data. 
While we used a linear data term in the node potential, $\abf^T\xbf_i$ in~\eqref{eq:phiN}, and a polynomial kernel of degree 1 in the pairwise potential to learn features from data, $\boldsymbol{\nu}^T(\xbf_i\xbf_j)$ in~\eqref{eq:phiE}, there are possibilities of using more complex data terms to learn more expressive features, like using a Gaussian kernel as in~\cite{LearningCRF2014Orlando}. Another interesting direction could be to use a smaller neural network to learn pairwise or higher-order features from the node features. On a GNU/Linux based standard computer with 32 GB of memory running one full cross validation procedure on 32 images upto 6 hours. Predictions using a trained MFN takes less than a minute per image.

Our model can be seen as an intermediate between an entirely model-based solution and an end-to-end learning approach. It can be interpreted as a structured neural network where the interactions between layers are based on the underlying graphical model, while the parameters of the model are learned from data. This, we believe, presents an interesting link between probabilistic graphical models and neural network-based learning. 

\subsection*{Acknowledgements}

This  work was funded by the Independent Research Fund Denmark (DFF) and Netherlands Organisation for Scientific Research (NWO).

\section{Appendix}
\label{sec:app}
We provide the proof for obtaining the mean field approximation update equations in~(6) and~(7) starting from the variational free energy in equation~(4). We start by repeating the expression for the node and pairwise potentials.

\subsubsection*{Node potential:}
\begin{equation}
	\phi_i(\sbf_i) = \sum_{v=0}^{2} \beta_v \Im \Big [ \sum_{j} s_{ij} = v\Big ] +  \abf^T \xbf_i\sum_{j} s_{ij},
	\label{eq:phiN_a}
\end{equation}
\subsubsection*{Pairwise potential:}
\begin{equation}
	\phi_{ij}(\sbf_i,\sbf_j) = \lambda \big( 1-2|s_{ij} - s_{ji}| \big ) + (2s_{ij}s_{ji}-1) \Big [ \boldsymbol{\eta}^T|\xbf_i-\xbf_j| + \boldsymbol{\nu}^T(\xbf_i\xbf_j)\Big]. 
	\label{eq:phiE_a}
\end{equation}

The variational free energy is given as,
\begin{equation}
 \Fcal(q_\Sbf) = \ln Z+ \Em_{q_{\Sbf}} \Big [ \ln p(\Sbf| \Xbf) - \ln q(\Sbf) \Big ].
 \label{eq:elbo_a}
\end{equation}
Plugging in~\eqref{eq:phiN_a} and~\eqref{eq:phiE_a} in~\eqref{eq:elbo_a}, we obtain the following:
\begin{align}
 \Fcal(q_\Sbf) &= \ln Z+ \Em_{q_{\Sbf}} \Big [ \sum_{i \in \Ncal} \Big\{ \beta_0 \Im \big [ \sum_{j} s_{ij} = 0\big ] + \beta_1 \Im \big [ \sum_{j} s_{ij} = 1\big ] + \beta_2 \Im \big [ \sum_{j} s_{ij} = 2\big ]+   \abf^T \xbf_i\sum_{j} s_{ij} \Big\}  \nonumber \\
&+ \sum_{(i,j) \in \Ecal} \Big\{\lambda \big( 1-2|s_{ij} - s_{ji}| \big )+ (2s_{ij}s_{ji}-1) \Big [ \boldsymbol{\eta}^T|\xbf_i-\xbf_j| + \boldsymbol{\nu}^T(\xbf_i\xbf_j)\Big]\Big\} - \ln q(\Sbf) \Big ].
\end{align}
We next take expectation $\Em_{q_{\Sbf}}$ using the mean-field factorisation that $q(\Sbf) = \prod_{i=1}^N \prod_{j\in \Ncal_i} q_{ij}(s_{ij})$ and the fact that $\Pr\{s_{ij}=1\} = \alpha_{ij}$ we simplify each of the factors :
\begin{align}
\Em_{q_{\Sbf}} \Big [  \beta_0 \Im \big [ \sum_{j} s_{ij} = 0\big ] \Big] = 
\Em_{q_{{i1}}\dots q_{{iN}}} \beta_0\Im \big [ \sum_{j} s_{ij} = 0\big ] \Big] = \beta_0 \prod_{j\in \Ncal_i}  (1-\alpha_{ij}).
\end{align}
Similarly,
\begin{align}
\Em_{q_{\Sbf}} \Big [  \beta_1 \Im \big [ \sum_{j} s_{ij} = 1\big ] \Big] =  \beta_1 \prod_{j\in \Ncal_i}  (1-\alpha_{ij}) \sum_{j\in \Ncal_i} \frac{\alpha_{im}}{ (1-\alpha_{im})}
\end{align}
and
\begin{align}
\Em_{q_{\Sbf}} \Big [  \beta_2 \Im \big [ \sum_{j} s_{ij} = 2\big ] \Big] =  \beta_2 \prod_{j\in \Ncal_i}  (1-\alpha_{ij}) \sum_{m\in \Ncal_i} \sum_{n \in \Ncal_i\backslash m} \frac{\alpha_{im}}{(1-\alpha_{im})}\frac{\alpha_{in}}{(1-\alpha_{in})}.
\end{align}
Next, we focus on the pairwise symmetry term:
\begin{equation}
\Em_{q_{\Sbf}} \Big [ \lambda \big( 1-2|s_{ij} - s_{ji}| \big ) \Big] =\lambda \big( 1-2(\alpha_{ij} +\alpha_{ji}) +4\alpha_{ij}\alpha_{ji}\big)
\end{equation}
Using these simplified terms, and taking the expectation over the remaining terms, we obtain the ELBO as,
\begin{align}	
	& \Fcal{q_{\Sbf}} =  \ln Z+ \sum_{i \in \Ncal} \prod_{j \in \Ncal_i} (1-\alpha_{ij}) \Big \{ \beta_0  
+ \sum_{m \in \Ncal_i} \frac{\alpha_{im}} {(1-\alpha_{im})} \Big[ \beta_1   + \beta_2 \sum_{n \in \Ncal_i \setminus m} \frac{ \alpha_{in}}{(1-\alpha_{in})} \Big] 
\nonumber \\
&+
\abf^T \xbf_i\sum_{j} \alpha_{ij} \Big\}
 + \sum_{i \in \Ncal} \sum_{j \in \Ncal_i} \Big\{ \lambda \big( 1-2(\alpha_{ij} +\alpha_{ji}) +4\alpha_{ij}\alpha_{ji}\big) -\Big( \alpha_{ij} \ln {\alpha_{ij}} \nonumber \\
 &  + (1-\alpha_{ij}) \ln (1-{\alpha_{ij}}) \Big) 
 + (2\alpha_{ij}\alpha_{ji}-1) \Big [ \boldsymbol{\eta}^T|\xbf_i-\xbf_j| + \boldsymbol{\nu}^T(\xbf_i\xbf_j)\Big]\Big \}.
	\label{eq:elbo_aw_a}
\end{align}

We next differentiate ELBO in~\eqref{eq:elbo_aw_a} wrt $\alpha_{kl}$ and set it to zero. 
\begin{align}
	\frac{\partial\Fcal{q_{\Sbf}}}{\partial \alpha_{kl}} &=   -\Big [\ln \frac{\alpha_{kl}}{1-\alpha_{kl}} \Big ] +
\prod_{j \in \Ncal_k \setminus l} \big(1-\alpha_{kj}\big) \Big\{ \sum_{m \in \Ncal_k \setminus l} \frac{\alpha_{km}}{(1-\alpha_{km})}
	\Big[ (\beta_2-\beta_1) - \beta_2 \sum_{n \in \Ncal_k \setminus l,m} \frac{\alpha_{kn}}{(1-\alpha_{kn})}\Big]
     \nonumber \\ 
    &+ \big(\beta_1-\beta_0 \big) \Big\} 
    + \abf^T \xbf_i + (4\alpha_{lk}-2)\lambda + 2\alpha_{lk}\big( \boldsymbol{\eta}^T|\xbf_i-\xbf_j| + \boldsymbol{\nu}^T(\xbf_i\xbf_j) \big).
	\qquad = 0
\end{align}
From this we obtain the MFA update equation for iteration $(t+1)$ based on the states from $(t)$,
\begin{equation}
	\mathlarger \alpha_{kl}^{(t+1)} = \mathlarger \sigma({\gamma_{kl}}) = \frac{1}{1+\exp^{-\gamma_{kl}}} \text{ } \forall \text{ } {k} = \{1\dots N\},\text{ } l \in \Ncal_k \text{ }:\text{ } |\Ncal_k|=L
    \label{eq:mfaUp_a}
\end{equation}
where $\mathlarger \sigma(.)$ is the sigmoid activation function, $\Ncal_k$ are the $L$ nearest neighbours of node $k$ based of positional Euclidean distance, and 
\begin{align}
	&\mathlarger \gamma_{kl} = 
\prod_{j \in \Ncal_k \setminus l} \big(1-\alpha_{kj}^{(t)}\big) \Big\{ \sum_{m \in \Ncal_k \setminus l} \frac{\alpha_{km}^{(t)}}{(1-\alpha_{km}^{(t)})}
	\Big[ (\beta_2-\beta_1) - \beta_2 \sum_{n \in \Ncal_k \setminus l,m} \frac{\alpha_{kn}^{(t)}}{(1-\alpha_{kn}^{(t)})}\Big]
     \nonumber \\ 
    &+ \big(\beta_1-\beta_0 \big) \Big\} 
    + \abf^T \xbf_i + (4\alpha_{lk}^{(t)}-2)\lambda + 2\alpha_{lk}^{(t)}\big( \boldsymbol{\eta}^T|\xbf_i-\xbf_j| + \boldsymbol{\nu}^T(\xbf_i\xbf_j) \big).
    \label{eq:mfa_a}
\end{align}

\end{document}